%% file: paper.tex
\documentclass[]{bytedance_seed}

\usepackage[toc,page,header]{appendix}

\usepackage{minitoc}

\title{Hand-in-the-Loop: Improving VLA Policies for Dexterous Manipulation via Seamless Hand-Arm Intervention}

\author[1,2,3,*]{Zhuohang Li}
\author[3]{Liqun Huang}
\author[3]{Wei Xu}
\author[3]{Zhengming Zhu}
\author[3,4,*]{Nie Lin}
\author[3]{Xiao Ma}
\author[1,2,\dagger]{Xinjun Sheng}
\author[3,\dagger]{Ruoshi Wen}

\affiliation[1]{State Key Laboratory of Mechanical System and Vibration, School of Mechanical Engineering, Shanghai Jiao Tong University}
\affiliation[2]{Shanghai Key Laboratory of Intelligent Robotics, Meta Robotics Institute, Shanghai Jiao Tong University, Shanghai 200240, China}
\affiliation[3]{ByteDance Seed}
\affiliation[4]{The University of Tokyo}

\contribution[*]{Work done at ByteDance Seed}
\contribution[\dagger]{Corresponding authors}

\abstract{
Vision-Language-Action (VLA) models are prone to compounding errors in dexterous manipulation, where high-dimensional action spaces and contact-rich dynamics amplify small policy deviations over long horizons. While Interactive Imitation Learning (IIL) can refine policies through human correction data, applying it to high-degree-of-freedom (DoF) robotic hands remains challenging due to a command mismatch between human teleoperation and policy execution at the intervention moment, which causes abrupt robot-hand configuration changes, or “gesture jumps”. We present Hand-in-the-Loop (HandITL), a seamless human-in-the-loop intervention method that blends human corrective intent with autonomous policy execution to avoid gesture jumps during bimanual dexterous manipulation. Compared with taking over control using direct teleoperation, HandITL reduces intervention jitter by 99.8\% and preserves robust post-intervention manipulation, reducing grasp failures by 87.5\% and mean completion time by 19.1\%. We validate HandITL on tasks requiring bimanual coordination, tool use, and fine-grained long-horizon manipulation. When used to collect correction data for policy refinement, HandITL yields policies that outperform those trained with standard teleoperation data by 19\% on average across three long-horizon dexterous tasks.
}

\date{\today}
\correspondence{Xinjun Sheng at \email{xjsheng@sjtu.edu.cn}, Ruoshi Wen at \email{wenruoshi@bytedance.com}}

\checkdata[Project Page]{\url{https://simpson-li.github.io/Hand-in-the-Loop/}}

\begin{document}
\maketitle


\section{Introduction}
\input{sections/intro}

\section{Related Work}
\input{sections/related_works}

\section{Methodology}
\input{sections/method}


\section{Experiments}
\input{sections/experiments}

\section{Conclusion and Limitations}
\input{sections/conclusion}

\clearpage

\bibliographystyle{plainnat}
\bibliography{main}

\clearpage



\end{document}

%% file: sections/intro.tex
\label{sec:intro}

Vision-Language-Action (VLA) models have demonstrated strong potential in robotic manipulation~\cite{zitkovich2023rt,kim2024openvla}. However, transferring these policies from parallel-jaw grippers to high-degree-of-freedom (DoF) anthropomorphic hands remains challenging due to the higher-dimensional action space and complex finger-object contacts, which make the system highly sensitive to small hand-pose errors. These factors amplify deployment-time distribution shift on real dexterous systems, where small deviations in fingertip motion, contact state, or object pose can quickly compound during contact-rich, long-horizon tasks, driving the system into out-of-distribution (OOD) states from which the base policy often fails to recover.~\cite{ross2011reduction,welte2025interactive}.
Interactive Imitation Learning (IIL), such as Dataset Aggregation (DAgger)~\cite{ross2011reduction,kelly2019hg}, mitigates this issue by enabling human intervention during policy rollouts, thereby collecting targeted recovery data from deployment-induced states and improving policy robustness beyond the original training distribution.

\begin{figure}[t]
    \centering
    \includegraphics[width=0.9\linewidth]{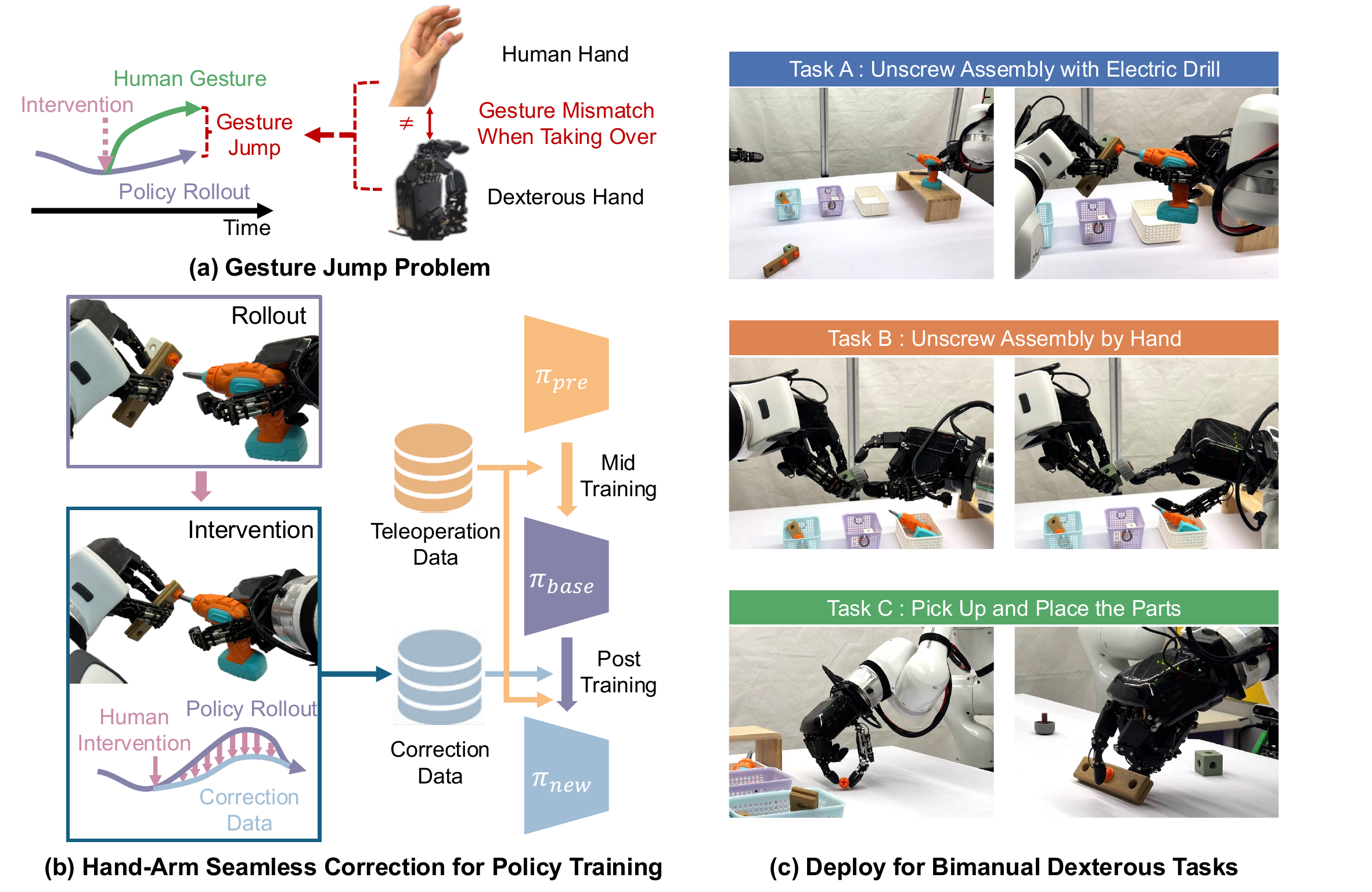}
    \caption{\textbf{Overview of HandITL.} 
    \textbf{(a)} Misalignment between the human hand and the robot hand at the intervention moment can induce a gesture jump. \textbf{(b)} HandITL maintains command continuity at intervention and enables smooth correction. The resulting correction data are then used for policy refinement. \textbf{(c)} The fine-tuned policy is deployed to perform complex, long-horizon bimanual dexterous tasks, such as unscrewing assemblies and precision part sorting.}
    \label{fig:overview}
\end{figure}

Despite its promise, applying IIL to dexterous manipulation remains challenging due to three key bottlenecks: intervention moment command discontinuities between human teleoperation and policy execution, limited hand-arm coordination under shared autonomy, and a limited understanding of how high-dimensional correction data improves policy learning. Low-DoF interfaces, such as joysticks or SpaceMouse devices, lack the input dimensionality required for coordinated multi-finger corrections, while high-fidelity hand teleoperation systems~\cite{wang2024dexcap,wen2025gr} are primarily designed for collecting full demonstrations rather than enabling seamless corrective interventions during policy rollouts. The absence of such seamless intervention becomes critical at the intervention moment: conventional retargeting-based teleoperation abruptly replaces the policy-generated hand command with a command derived from the operator's current hand pose. If the robot is maintaining a stable grasp while the operator's hand is at rest or not aligned with the robot's current hand configuration, the resulting mismatch induces a large command discontinuity, which we refer to as a \textit{gesture jump}, as illustrated in Fig.~\ref{fig:overview}(a). Such jumps can destabilize ongoing dexterous grasps and cause object drops, often making recovery difficult or impossible within the same rollout.

To overcome these limitations, we propose Hand-in-the-Loop (HandITL), a seamless intervention method for bimanual dexterous VLA policies, as illustrated in Fig.~\ref{fig:overview}. HandITL fuses human corrective intent with the policy's ongoing action stream, reducing intervention-induced command discontinuities while preserving ongoing contact-rich manipulation. For the hands, HandITL performs optimization-based relative retargeting anchored at the intervention moment, transferring the operator's incremental fingertip motions rather than matching the absolute human hand pose. This enables smooth multi-finger corrections under grasp-preserving, kinematic safety, and velocity constraints. For the arms, HandITL uses a velocity-based shared-control interface that injects transient wrist motions as residual twists into policy-predicted arm commands, enabling localized corrections while mitigating long-term drift.

HandITL supports two intervention modes for long-horizon bimanual manipulation: full takeover for critical recovery and copilot shared-control for small residual corrections. We evaluate these modes on real-world tasks involving bimanual coordination, tool use, and fine-grained manipulation. Extensive experiments show that HandITL reduces intervention-induced command discontinuity by 99.8\%. Moreover, fine-tuning the base VLA policy with the collected \textit{on-policy} correction data substantially improves long-horizon task performance, outperforming teleoperation-data fine-tuning under the same training budget by 19\% on average across three tasks. The main contributions of this work are summarized as follows:

\begin{itemize}
    \item \textbf{A Seamless Intervention Method for Dexterous VLA policies.} We introduce HandITL, a high-DoF human-in-the-loop intervention method that enables online corrective data collection during long-horizon, contact-rich bimanual dexterous manipulation.

    \item \textbf{Optimization-Based Relative Hand Retargeting.} We develop a retargeting algorithm that transfers the operator's relative fingertip motions, rather than absolute hand poses, to the robot hand. This relative formulation substantially reduces command discontinuities at the intervention moment and preserves ongoing contacts after intervention.

    \item \textbf{Velocity-Based Arm Shared Control.} We design a bimanual arm shared-control interface that injects human corrective intent as transient residual twists into policy-predicted arm actions, enabling localized corrections without disrupting the policy's active rollout.

    \item \textbf{On-Policy Intervention Data for Policy Refinement.} We empirically demonstrate that high-dimensional, on-policy correction data collected through HandITL improves long-horizon dexterous manipulation performance after fine-tuning and reduces failures caused by compounding errors.
\end{itemize}

%% file: sections/related_works.tex
\subsection{VLA Models for Dexterous Manipulation}

Recent VLA models have shown promising instruction-conditioned manipulation capabilities through large-scale pre-training~\cite{zitkovich2023rt,kim2024openvla,black2024pi_0,intelligence2025pi_05,intelligence2025pi,cheang2025gr,li2025gr}. Yet, most successful deployments still focus on relatively low-DoF end effectors, and extending these models to multi-fingered anthropomorphic hands remains challenging. Existing dexterous VLA systems broadly follow two paradigms. Hierarchical approaches decouple semantic planning from low-level dexterous control~\cite{zhong2026dexgraspvla,yuan2025being,mao2024dexskills}, improving modularity but potentially limiting closed-loop adaptation when contact states, object poses, or grasp configurations deviate from the planned trajectory. End-to-end approaches instead learn direct mappings from multimodal observations and instructions to high-DoF actions~\cite{wen2025gr,luo2025being,bjorck2025gr00t}, but require substantial real-world dexterous data and remain vulnerable to compounding errors during long-horizon execution. Our work focuses on how deployed end-to-end VLA policies for dexterous manipulation can be improved using on-policy correction data collected from real bimanual policy rollouts.

\subsection{Interactive Imitation Learning for Policy Fine-Tuning}

IIL mitigates deployment-time distribution shift by collecting expert supervision on states induced by the learner's own policy. Classical DAgger and human-gated variants reduce covariate shift through expert queries or interventions during rollouts~\cite{ross2011reduction,kelly2019hg}, and recent extensions improve sample efficiency, compliance, or intervention-aware policy refinement~\cite{bicer2019sample,liu2025takead,xu2025compliant,lee2025diff}. This paradigm has been effective in relatively low-dimensional manipulation settings, where corrective labels can be provided through low-dimensional control interfaces and incorporated as on-policy recovery demonstrations.

Its extension to bimanual dexterous VLA policies, however, remains underexplored. From a data perspective, high-dimensional correction data may provide valuable recovery supervision, but it may also introduce noisy, sparse, or distributionally distinct labels that interfere with previously learned manipulation skills. From a systems perspective, studying this question requires a real-time interface capable of collecting physically consistent hand-arm corrections without causing command discontinuities or contact disruption. Existing IIL pipelines do not directly address this interface requirement for high-DoF dexterous hands. HandITL fills this gap by enabling seamless online intervention and on-policy corrective data collection during real bimanual dexterous rollouts.

\subsection{Interfaces for Real-Time Human Correction}

The effectiveness of IIL depends on interfaces that allow operators to provide timely and physically consistent corrections during policy rollouts. Low-dimensional devices, such as joysticks~\cite{spencer2020learning,hu2025rac}, SpaceMouse controllers~\cite{luo2025precise,chen2025conrft}, or smartphones~\cite{mandlekar2020human}, are effective for correcting simple robot trajectories but cannot naturally express coordinated finger-level corrections. High-fidelity teleoperation systems, including hand/arm exoskeletons~\cite{forouhar2024tactile,zhong2025nuexo,wei2024wearable,zhang2025doglove}, provide richer control but require specialized hardware and calibration. Glove-based retargeting pipelines~\cite{handa2020dexpilot,li2022dexterous,wang2024dexcap,wen2025gr} provide an accessible alternative of collecting teleoperation data, but they are not designed for collecting human correction data during policy rollouts.

This distinction becomes critical in dexterous IIL. Directly switching from policy-generated commands to absolute pose retargeting can produce a gesture jump when the operator's hand pose is not aligned with the current robot hand pose, destabilizing contact-rich manipulation. Recent systems such as CR-DAgger~\cite{xu2025compliant} and RoboCopilot~\cite{wu2025robocopilot} explore shared-control interfaces that blend human corrections with policy rollouts. However, their extension to high-DoF bimanual dexterous manipulation remains nontrivial, as hand-arm corrections must be injected without disrupting ongoing contacts or introducing intervention discontinuities. HandITL addresses this gap through relative hand retargeting and velocity-based arm shared control, enabling physically consistent online correction and on-policy data collection during real dexterous rollouts.

%% file: sections/method.tex
\label{sec:method}
\begin{figure}
\centering
\includegraphics[width=1.0\linewidth]{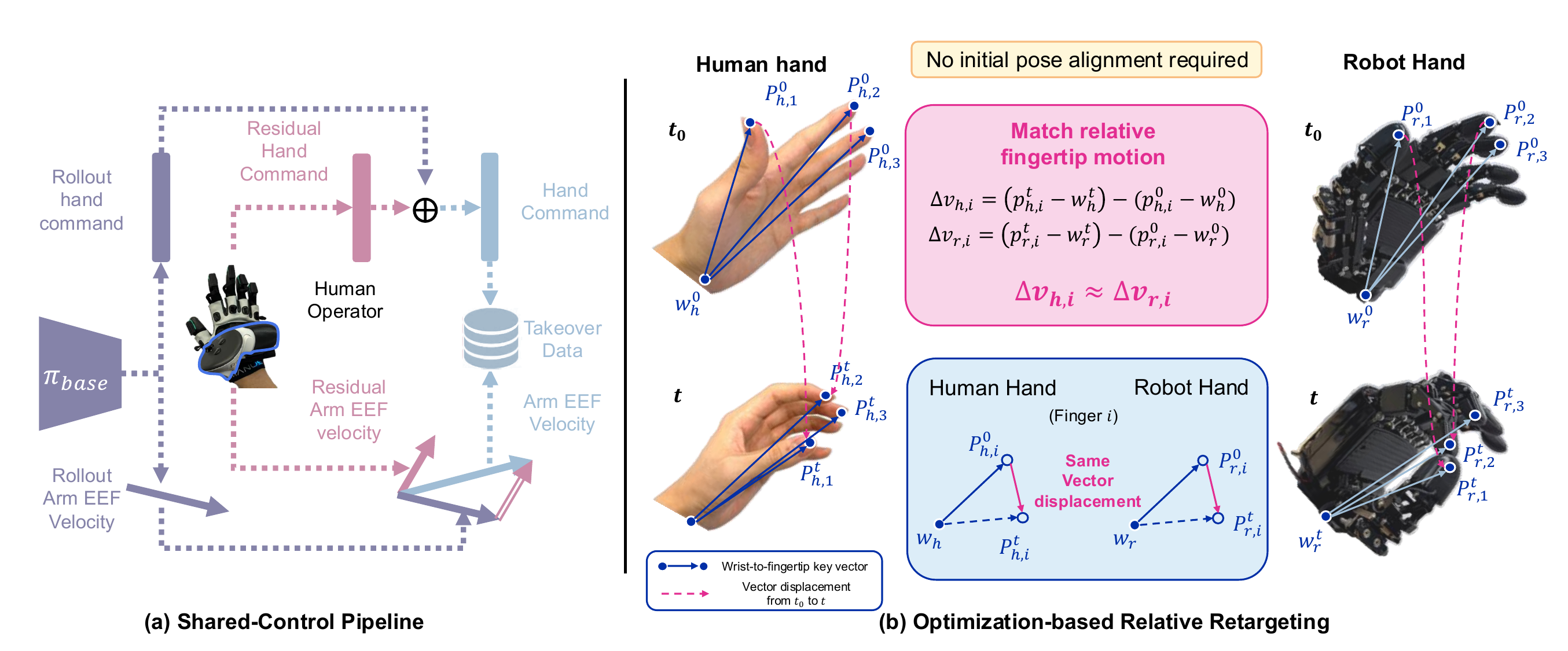}
\caption{\textbf{Architecture of the seamless interventional method.} \textbf{(a)} The base VLA policy ($\pi_{base}$) generates autonomous rollout commands. During an intervention, the human operator injects transient residual hand commands and velocity-based arm twists. These corrective signals are seamlessly blended with the rollout commands to form the final executed actions. \textbf{(b)} Our method anchors at the intervention moment $t_0$ and maps the human's relative motion to the robot. It preserves existing contacts at the intervention moment because no robot hand motion will be caused with no human hand motion.}
\label{fig:method}
\end{figure}

\subsection{HandITL Pipeline}
\label{subsec:system_overview}

HandITL is a real-time intervention method that enables high-dimensional human corrections during ongoing dexterous VLA rollouts. During autonomous execution, the VLA policy predicts hand-arm commands from visual observations and proprioception. When the operator anticipates a potential failure, they trigger intervention and provide corrective hand and wrist commands. As illustrated in Fig.~\ref{fig:method}(a), HandITL decomposes the operator's corrections into two streams. For the dexterous hands, optimization-based relative retargeting maps the operator's incremental fingertip motions from intervention onset to robot hand commands while preserving the current grasp state, as shown in Fig.~\ref{fig:method}(b). For the arms, velocity-based shared control converts transient wrist motions into residual end-effector twists and injects them into the policy-predicted arm commands, enabling corrections without persistent drift.
HandITL supports both full takeover for substantial recovery and copilot shared control for small residual corrections, producing on-policy rollout trajectories that combine autonomous execution with human-corrected segments. The resulting hand-arm commands are executed on the robot and recorded with the corresponding observations.

\subsection{Problem Formulation and Intervention Modes}
\label{subsec:problem_formulation}

We formulate bimanual dexterous manipulation as a sequential decision-making process governed by a VLA policy $\pi_\theta$. At each time step $t$, the policy receives an observation $o_t$, including multi-view RGB-D images and proprioception, and predicts an action sequence $\mathbf{A}_t^\pi=\{\mathbf{a}_{t:t+H-1}^{\pi}\}$ over a horizon $H$. During autonomous rollout, the robot executes the first action $\mathbf{a}_t^\pi$ in a receding-horizon manner.

During an intervention, the operator provides human corrective input $\mathbf{u}_t^h$, which is fused with the policy command to generate the executed command:
$
\mathbf{a}_{t}^{exec} =
\left(
\mathcal{F}_a(\mathbf{a}_{t,a}^{\pi}, \mathbf{u}_{t,a}^{h}; \beta_a),
\mathcal{F}_h(\mathbf{a}_{t,h}^{\pi}, \mathbf{u}_{t,h}^{h}; \beta_h)
\right),
$
where $\mathcal{F}_a$ denotes arm-level residual shared control, $\mathcal{F}_h$ denotes hand-level relative retargeting, and $\beta_a,\beta_h$ are the corresponding human authority weights. HandITL supports two intervention modes: \textbf{full takeover}, where human authority dominates for substantial recovery, and \textbf{copilot shared-control}, where the policy remains primary while human inputs provide local residual corrections.

The resulting intervention segments are aggregated into an on-policy correction dataset
$
\mathcal{D}_{\text{corr}}=
\{(o_t,\mathbf{a}_{t}^{exec})\}_{t\in\mathcal{I}},
$
where $\mathcal{I}$ denotes timesteps under human intervention. These physically executed correction labels are used to fine-tune the base policy $\pi_{\text{base}}$.

\subsection{Optimization-Based Relative Hand Retargeting}
\label{subsec:hand_retargeting}

Conventional pose-based retargeting directly maps the operator's absolute hand pose to the robot hand. During intervention, this can induce gesture jumps when the operator's hand configuration is not aligned with the current robot hand configuration. To reduce this discontinuity, we propose an \textbf{optimization-based relative retargeting} method anchored at the intervention timestamp $t_0$.
We define two types of hand key vectors: wrist-to-fingertip vectors $\mathbf{v}_i$, which describe global finger motion, and thumb-to-fingertip opposition vectors $\mathbf{u}_j$, which describe pinch-related motion. Instead of matching absolute human key vectors, our method tracks their changes after intervention:

\begin{equation}
    \Delta \mathbf{v}_{i}^{rob}(q_t)=\mathbf{v}_{i}^{rob}(q_t)-\mathbf{v}_{i}^{rob}(q_{t_0}), \quad
    \Delta \hat{\mathbf{v}}_{i}^{hum}=\hat{\mathbf{v}}_{i}^{hum}(t)-\hat{\mathbf{v}}_{i}^{hum}(t_0)
\end{equation}
where $\hat{(\cdot)}$ denotes human key vectors transformed into the robot hand frame with scale normalization, as illustrated in Fig.~\ref{fig:method}(b). At each control step, we solve for the robot hand configuration $q_t$ by minimizing

\begin{equation}
\label{eq:hand_opt}
\begin{aligned}
\mathcal{J}(q_t) = 
& \underbrace{\sum_{i=1}^{5}
\beta_i(d_i)
\mathcal{H}_{\delta}
\left(
\left\|
\Delta \mathbf{v}_{i}^{rob}(q_t)
-
\Delta \hat{\mathbf{v}}_{i}^{hum}
\right\|_2
\right)}_{\mathcal{L}_{\text{shape}} \text{ (Global Shaping)}} 
 + \underbrace{\sum_{j\in\mathcal{F}_{opp}}
\omega_j(d_j)
\mathcal{H}_{\delta}
\left(
\left\|
\mathbf{u}_{j}^{rob}(q_t)
-
\alpha_j(d_j)\mathbf{u}_{j}^{tgt}(d_j)
\right\|_2
\right)}_{\mathcal{L}_{\text{grasp}} \text{ (Precision Grasping)}} \\
& + \underbrace{
\gamma
\sum_{(m,n)\in\mathcal{C}}
\left[
\max(0,d_{\text{safe}}-D_{mn}(q_t))
\right]^2}_{\mathcal{L}_{\text{safe}} \text{ (Structural Safety)}} 
 + \underbrace{
\lambda_{\text{reg}}
\mathcal{H}_{\delta}
\left(
\left\|
q_t-q_{t-1}
\right\|_2
\right)}_{\mathcal{L}_{\text{reg}} \text{ (Temporal Regularization)}}.
\end{aligned}
\end{equation}
The loss components are defined as follows:

\begin{itemize}
    \item \textbf{Global Shaping ($\mathcal{L}_{\text{shape}}$):} $\beta_i(d_i)$ is a distance-dependent gate based on the human thumb-to-finger distance $d_i$. This term dominates when the hand is open and gradually decreases as the human hand enters a pinch configuration.

    \item \textbf{Precision Grasping ($\mathcal{L}_{\text{grasp}}$):} $\mathcal{F}_{opp}$ denotes the set of fingertips paired with the thumb for opposition-based grasping. For pinch grasping, we define a nominal target opposition vector ${\mathbf{u}}_{j}^{\text{tgt}} = \mathbf{u}_{j}^{\text{rob}}(q_{t_0}) + \Delta \hat{\mathbf{u}}_{j}^{\text{hum}}$ and a pinch activation factor $\alpha_j(d_j) \in [0,1]$. When $d_j$ falls below a threshold, $\alpha_j(d_j) \to 0$, biasing the opposition vector toward a compact pinch configuration. As $d_j$ decreases, $\omega_j(d_j)$ increases, causing the optimization to prioritize pinch consistency.

    \item \textbf{Structural Safety ($\mathcal{L}_{\text{safe}}$):} $D_{mn}(q_t)$ is the physical distance between potentially colliding hand links or joints. The hinge penalty activates only when $D_{mn}$ falls below the safety margin $d_{\text{safe}}$, with $\gamma$ controlling its strength.

    \item \textbf{Temporal Regularization ($\mathcal{L}_{\text{reg}}$):} Penalizes the joint displacement, scaled by $\lambda_{\text{reg}}$, to prevent high-frequency jitter.

\end{itemize}

\subsection{Velocity-Based Shared Arm Control}
\label{subsec:arm_control}

Because the operator's hands are occupied by the glove-mounted VR controllers, conventional joystick-based correction is impractical. To enable intuitive corrections, we design a \textbf{velocity-based shared-control} interface. Human wrist motions are translated into residual twists, smoothed via Exponential Moving Average (EMA), and injected into the policy-predicted arm command without accumulating persistent drift.

At step $t$, the VLA outputs the target pose and feedforward twist $(\mathbf{p}_t^{\pi}, \mathbf{R}_t^{\pi}, \dot{\mathbf{p}}_t^{\pi}, \boldsymbol{\omega}_t^{\pi})$. Simultaneously, the residual human twist $(\dot{\mathbf{p}}_t^{h}, \boldsymbol{\omega}_t^{h})$ is estimated over a window of $k=2$ VR ticks ($\Delta T=0.04$\,s) to balance stability and latency:
\begin{equation}
    \dot{\mathbf{p}}_{t}^{h} = \text{EMA}\left(\frac{\mathbf{p}_t^{vr} - \mathbf{p}_{t-k}^{vr}}{\Delta T}\right), \quad
    \boldsymbol{\omega}_{t}^{h} = \text{EMA}\left(\frac{\log\left((\mathbf{R}_{t-k}^{vr})^{-1}\mathbf{R}_t^{vr}\right)}{\Delta T}\right),
\end{equation}
where $\log(\cdot)$ denotes the logarithm map from $SO(3)$ to its axis-angle vector. The resulting residual twists are transformed into the robot base frame before fusion.
These twists are integrated as 
$\Delta \mathbf{p}_t^{h}=g_p\dot{\mathbf{p}}_t^{h}\Delta t$ and 
$\Delta \mathbf{R}_t^{h}=\exp(g_R\boldsymbol{\omega}_t^{h}\Delta t)$,
and then composed with the policy target:
\begin{equation}
    \mathbf{p}_t^{\text{tgt}} = \mathbf{p}_t^{\pi} + \Delta \mathbf{p}_t^{h}, \quad 
    \mathbf{R}_t^{\text{tgt}} = \mathbf{R}_t^{\pi} \Delta \mathbf{R}_t^{h}.
\end{equation}
The commanded spatial velocity is computed by a task-space PD tracker.
Crucially, deriving residuals from relative motion ensures that when the operator is still, $(\dot{\mathbf{p}}_t^{h}, \boldsymbol{\omega}_t^{h}) \rightarrow \mathbf{0}$. The EMA filter smoothly decays these residuals, preventing persistent offsets from accumulating without requiring the operator to manually return to a neutral ``center'' position.

%% file: sections/experiments.tex
\label{sec:experiments}

We aim to answer the following research questions through experiments:

\begin{itemize}
    \item \textbf{Q1:} Does our optimization-based relative retargeting method effectively reduce \textit{command discontinuity} (i.e., gesture jumps) at the moment of intervention?
    \item \textbf{Q2:} After intervention, does our human-in-the-loop interface preserve dexterous manipulation capability, such as maintaining grasps and completing fine-grained corrective motions?
    \item \textbf{Q3:} Can on-policy correction data collected via our method, especially under \textit{copilot shared-control}, improve VLA policy learning more effectively than additional teleoperation data on challenging long-horizon bimanual dexterous tasks?
\end{itemize}

\subsection{Experiments Overview}
\label{subsec:overview}

Our experiments consist of three parts corresponding to Q1-Q3: intervention command discontinuity analysis, post-intervention manipulation evaluation, and policy evaluation on long-horizon tasks. The first two evaluate whether HandITL enables stable and dexterous real-time human correction, while the third evaluates whether the collected on-policy correction data improves VLA policy learning.

For the first two parts, we compare our optimization-based relative retargeting method with three intervention method baselines: \textbf{Jacobian-based mapping}, which computes robot joint increments from fingertip displacements using the hand Jacobian~\cite{wang2024dexcap}; \textbf{Delta-command retargeting}, which applies relative command changes from an absolute teleoperation backend, $q_t^{exec}=q_{t_0}^{rob}+(q_t^{tel}-q_{t_0}^{tel})$~\cite{wen2025dexterous,handa2020dexpilot}; and \textbf{Direct teleoperation switching}, which directly switches to absolute teleoperation retargeting at the intervention moment~\cite{wen2025dexterous}.

For policy evaluation on long-horizon tasks, we deploy the base policy on real robots and record full on-policy rollouts containing both autonomous execution and human interventions. These correction rollouts are used for post-training and compared with equal-duration teleoperation data under the same training budget. The resulting policies are evaluated on three long-horizon bimanual dexterous manipulation tasks.

Experiments are conducted on a 56-DoF bimanual platform consisting of two 7-DoF Franka FR3 arms and two 21-DoF ByteDexter V2 hands~\cite{wen2025dexterous}. We collect real-world robot data using a bimanual teleoperation interface consisting of a Meta Quest VR setup for wrist-pose tracking, two Manus Metagloves for finger-motion capture, and foot pedals for triggering intervention. Two Meta Quest controllers are mounted on the dorsal side of the gloves to improve the reliability of coordinated wrist--hand tracking. The VLA policy receives multi-view RGB-D observations and robot proprioception.

\begin{figure}
    \centering
    \includegraphics[width=0.8\linewidth]{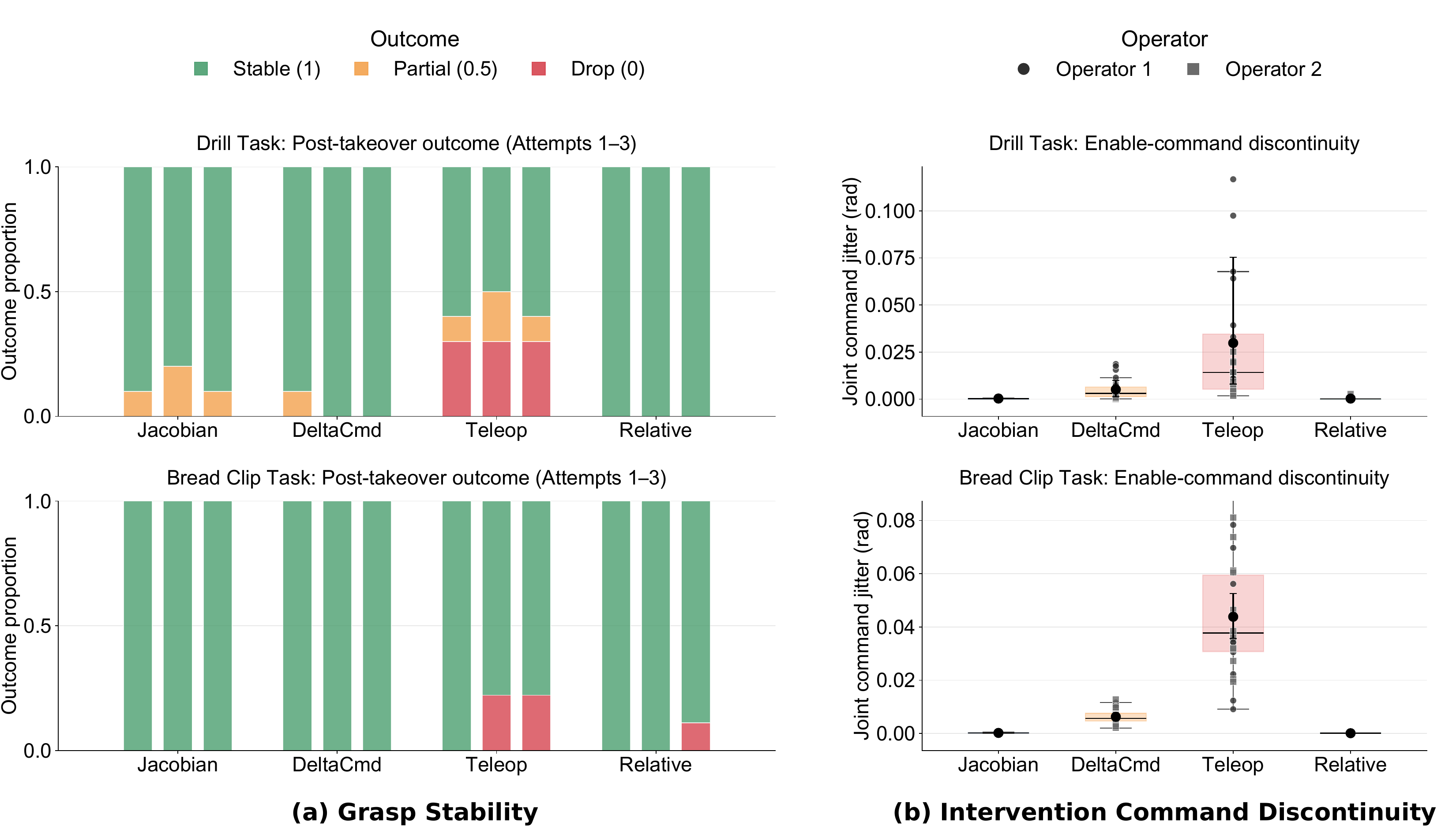}
    \caption{\textbf{Intervention command discontinuity on the Drill (top) and Bread Clip (bottom) tasks.}
    \textbf{(a)} Outcomes across three consecutive intervention attempts (10 trials each). Green, orange, and red denote stable grasps, partial failures (trigger loosening), and tool drops, respectively.
    \textbf{(b)} Attempt-level distribution of the mean command change at the intervention moment. Scatter points show individual interventions from two operators, with 95\% CI error bars.}
    \label{fig:jitter}
\end{figure}

\subsection{Intervention Command Discontinuity Analysis}

We evaluate intervention command discontinuity on the Drill and Bread Clip tasks. Intervention command discontinuity is measured as the delta hand joint command at the intervention moment,
$\|\mathbf{q}_{t_0^+}^{exec}-\mathbf{q}_{t_0^-}^{exec}\|_2$,
where $t_0^-$ and $t_0^+$ denote the control steps immediately before and after intervention. In each trial, the operator first establishes a stable grasp via teleoperation, switches to autonomous rollout, and toggles intervention three times during execution.

As illustrated in Fig.~\ref{fig:jitter}(a), across 30 trials per task-method pair, direct teleoperation switching frequently caused failures such as tool drops or drill-trigger release due to command discontinuity. In contrast, our relative retargeting method maintained grasp stability in nearly all trials; the only Bread Clip failure occurred after a successful handover due to subsequent operator manipulation rather than an intervention command discontinuity.

Our method substantially reduces command discontinuity, as shown in Fig.~\ref{fig:jitter}(b). On Bread Clip, where the open-hand posture amplifies human-robot mismatch, direct switching produces a large command jump (mean $\approx 4.38 \times 10^{-2}$). Our method reduces it to $\approx 6.8 \times 10^{-5}$, corresponding to a \textbf{99.8\%} reduction and nearly two orders of magnitude lower than DeltaCmd ($\approx 6.23 \times 10^{-3}$). On Drill, the closed power grasp yields smaller direct-switching jumps (mean $\approx 2.75 \times 10^{-2}$), but these still disrupt trigger actuation. Our method reduces the mean command change to $\approx 2.65 \times 10^{-4}$, comparable to the Jacobian baseline while avoiding explicit Jacobian inversion, thereby maintaining stable functional tool use.

\subsection{Post-Intervention Manipulation Capability}

We evaluate whether each intervention method preserves dexterous manipulation capability after control is transferred to the operator on the \textbf{Pick Up and Place the Parts} and \textbf{Pick Up the Drill} tasks in Fig.~\ref{fig:tasks}. Operators start directly in the active intervention state, isolating post-intervention manipulation from the handover discontinuity analyzed above. When reaching their physical workspace limit, operators can disengage, reset their hand posture, and re-engage, which we record as \textit{workspace resets}.

\input{sections/ability_ppa_drill_tables_test}

\begin{figure}
    \centering
    \includegraphics[width=0.6\linewidth]{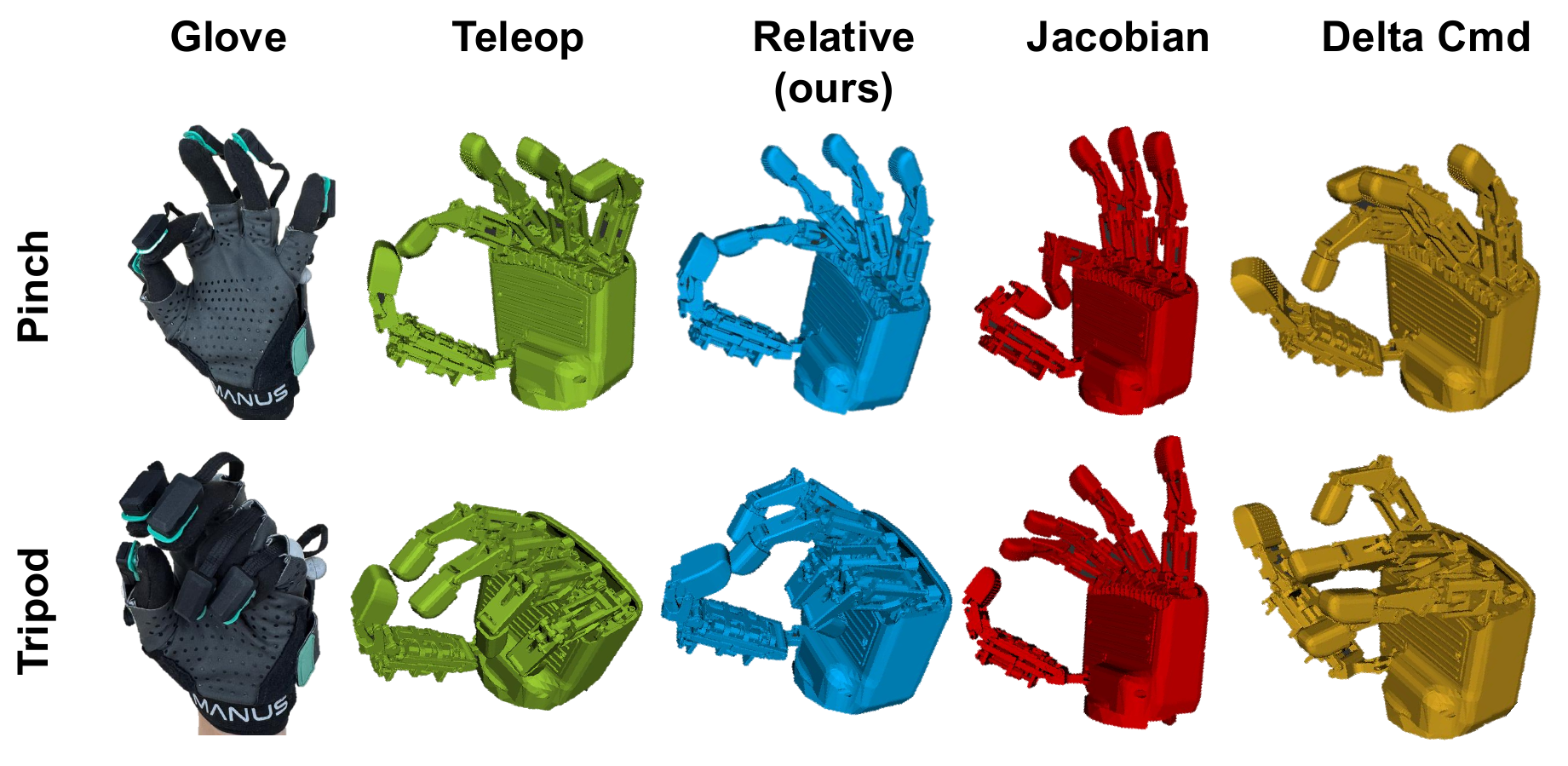}
    \caption{\textbf{Grasping Postures and Failure Modes.} Optimization-based methods (Teleop and our Relative approach) maintain precise thumb-to-finger opposition and avoid collisions. In contrast, differential methods (Jacobian and DeltaCmd) struggle with pinch grasping and suffer from accumulated drift during fast motions, leading to awkward postures and frequent grasp failures.}
    \label{fig:failure}
\end{figure}

As shown in Table.~\ref{tab:ability_ppa_drill}, on the Pick Up and Place the Parts task, our method achieves the lowest mean completion time ($42.8$\,s), reducing time by $19.1\%$ and grasp failures by $87.5\%$ compared to Teleop. It also shows a smaller cross-operator completion-time gap ($5.6$\,s vs. $18.0$\,s for Teleop), suggesting more consistent usability. As illustrated in Fig.~\ref{fig:failure}, the performance gap mainly stems from grasp-posture quality. Optimization-based methods, including Teleop and our Relative method, preserve more precise thumb-to-finger opposition for pinch and tripod grasps. In contrast, Jacobian and DeltaCmd rely on local differential updates without explicit grasp-geometry or structural-safety constraints, making them prone to fingertip misalignment, accumulated drift, and awkward or self-colliding postures during fast motions. These failures make object grasping more difficult, resulting in part drops and longer completion times ($68.0$\,s and $56.7$\,s, respectively).

On the Pick Up the Drill task, our method again achieves the lowest mean completion time ($14.4$\,s), the fewest workspace resets, and more consistent per-operator completion times than the differential baselines. Its fine-finger mapping differs slightly from absolute teleoperation, so operators sometimes require brief adaptation to precisely actuate the trigger without tactile feedback, slightly reducing button-triggering success. Overall, these results indicate that our method preserves efficient post-intervention dexterous manipulation while reducing grasp failures and workspace resets.

\subsection{Policy Evaluation on Long-Horizon Tasks}

\begin{figure}
    \centering
    \includegraphics[width=1.0\linewidth, page=2]{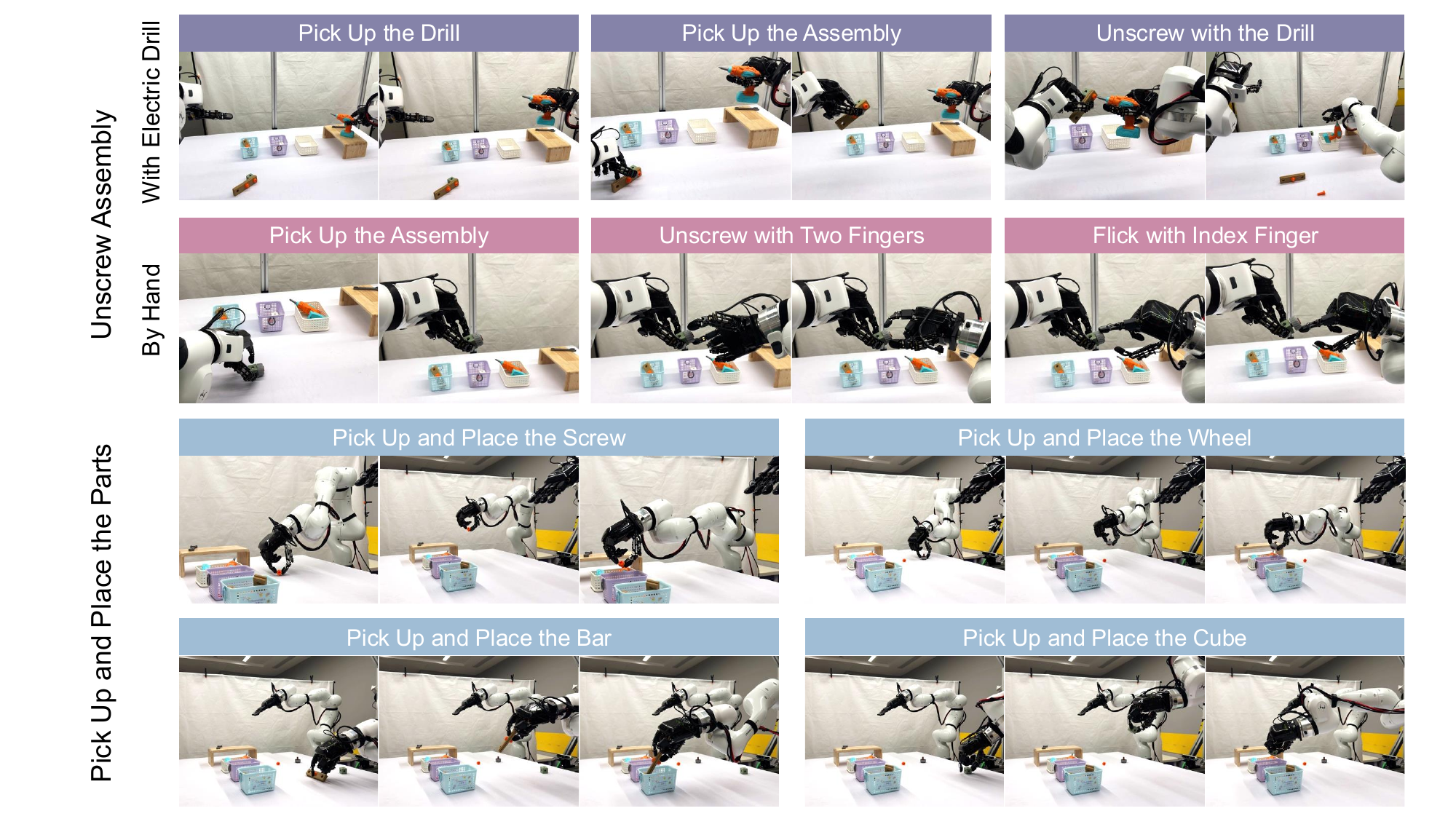}
    \caption{\textbf{Execution sequences of the three long-horizon bimanual dexterous manipulation tasks.} Top: \textit{Unscrew Assembly With Electric Drill}, requiring tool use and bimanual coordination. Middle: \textit{Unscrew Assembly By Hand}, demanding fine-grained in-hand manipulation. Bottom: \textit{Pick Up and Place the Parts}, involving sequential precision grasping of diverse small objects.}
    \label{fig:tasks}
\end{figure}

We collect a 20-hour real-world teleoperation dataset, evenly distributed across three long-horizon tasks. Starting from a pre-trained Gr-Dexter model~\cite{wen2025gr}, we fine-tune the model on this dataset to obtain the \textit{base policy}. We then deploy the base policy on real robots and collect correction data whenever the operator judges intervention to be necessary. Correction data are collected under both full-takeover and copilot shared-control modes in an HG-DAgger-style process~\cite{kelly2019hg}, where the operator intervenes during policy rollouts to correct failure-prone states. For both modes, we record the entire rollout, including autonomous and intervention segments.

For post-training, we compare the following policies:
\begin{itemize}
    \item \textbf{Base}: the policy obtained by fine-tuning the pre-trained VLA on the 20-hour teleoperation dataset;
    \item \textbf{Teleop\_old}: continuing to fine-tune the base policy on the original base dataset for the same number of post-training steps;
    \item \textbf{Teleop\_new}: fine-tuning the base policy using 1-hour new teleoperation data mixed with the base dataset;
    \item \textbf{Copilot}: fine-tuning the base policy using 1-hour copilot shared-control data mixed with the base dataset;
    \item \textbf{Full Takeover}: fine-tuning the base policy using 1-hour full-takeover data mixed with the base dataset.
\end{itemize}
For a fair comparison, all additional datasets are 1 hour long and are mixed with the base dataset at the same additional-to-base sampling ratio of 0.5:1 during post-training. All runs use identical training schedules and hyperparameters.

We evaluate all policies on three long-horizon tasks requiring high-precision bimanual coordination (Fig.~\ref{fig:tasks}). We report the average Sub-goal Completion Score over 10 evaluation rollouts for each model. As shown in Fig.~\ref{fig:tasks}, each long-horizon task is decomposed into 3-4 sub-tasks, each worth one point. Each sub-task is further divided into two or three ordered sub-stages, and the one-point score is evenly assigned to its sub-stages. The final score is computed as the sum of successfully completed sub-stage scores.

\begin{figure}
    \centering
    \includegraphics[width=1.0\linewidth]{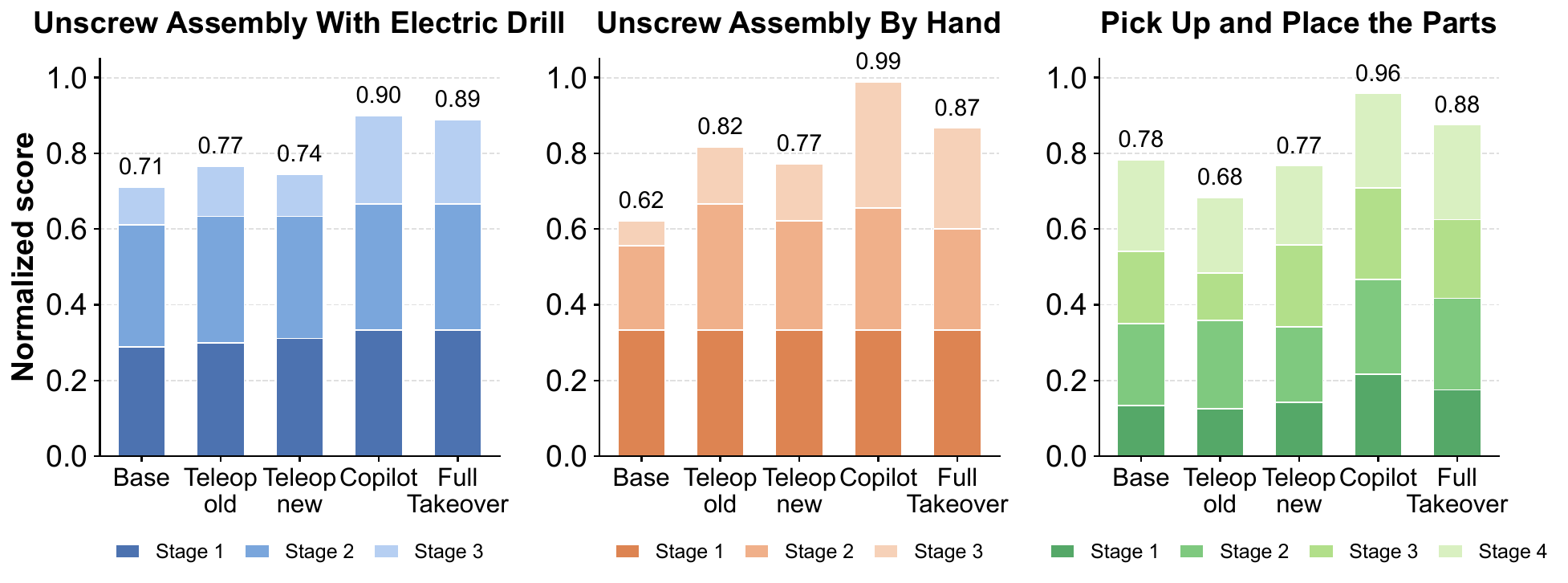}
    \caption{\textbf{Average normalized sub-goal completion scores across three long-horizon tasks.} Intervention-trained policies (\texttt{Copilot}, \texttt{Full Takeover}) outperform pure teleoperation baselines, indicating improved robustness against compounding errors.}
    \label{fig:model_ability}
\end{figure}

As shown in Fig.~\ref{fig:model_ability}, comparing the five policies leads to three main observations:

\textbf{1) Limited Improvements from Pure Teleoperation:} 
Simply increasing pure teleoperation data brings limited and inconsistent gains. Both \texttt{Teleop\_old} and \texttt{Teleop\_new} show only marginal improvements, with effects varying across tasks. This suggests that additional off-policy demonstrations may not adequately cover policy-induced states where compounding errors occur, especially in late phases requiring precise contact-rich manipulation.

\textbf{2) Effectiveness of Intervention Data:} 
Policies fine-tuned with intervention data achieve higher average normalized completion scores. Because these datasets are recorded from deployed policy rollouts and include human corrections at failure-prone moments, they provide supervision on policy-induced states and targeted recovery behaviors. This makes intervention data more effective than standard teleoperation demonstrations for improving long-horizon robustness.

\textbf{3) Copilot vs. Full Takeover:} 
Among the two intervention strategies, \texttt{Copilot} generally yields the strongest overall performance. Unlike \texttt{Full Takeover}, where human commands dominate the executed trajectory, \texttt{Copilot} keeps the policy as the primary controller while injecting local residual corrections. The resulting data stays closer to the policy's rollout distribution, leading to more stable downstream improvements.

%% file: sections/ability_ppa_drill_tables_test.tex
\begin{table*}[t]
    \vspace{2mm}
    \centering
    \caption{Post-intervention manipulation capability on the \textbf{Pick Up and Place the Parts} and \textbf{Pick Up the Drill} tasks. \textbf{Interv.} (Interventions) and \textbf{Retries} are reported as the total count across all 10 trials (5 per operator). \textbf{Time (Op1/Op2)} denotes the separate mean completion times for Operator 1 and Operator 2 to demonstrate cross-operator robustness. \textbf{Success} rate indicates whether the drill trigger was successfully actuated.}
    \label{tab:ability_ppa_drill}
    \small
    \setlength{\tabcolsep}{4.5pt}
    \renewcommand{\arraystretch}{1.15}
    \resizebox{\textwidth}{!}{
    \begin{tabular}{lcccccccc}
    \toprule
    \multirow{3}{*}{\textbf{Method}} & \multicolumn{4}{c}{\textbf{Task 1: Pick Up and Place the Parts}} & \multicolumn{4}{c}{\textbf{Task 2: Pick Up the Drill}} \\
    \cmidrule(lr){2-5} \cmidrule(lr){6-9}
    & \textbf{Time (s) $\downarrow$} & \textbf{Time (Op1/Op2) $\downarrow$} & \textbf{Interv. $\downarrow$} & \textbf{Retries $\downarrow$} & \textbf{Time (s) $\downarrow$} & \textbf{Time (Op1/Op2) $\downarrow$} & \textbf{Interv. $\downarrow$} & \textbf{Success $\uparrow$} \\
    \midrule
    Jacobian & $68.0 \pm 10.8$ & $66.5$ / $69.5$ & $31/10$ & $14/10$ & $19.4 \pm 5.6$ & $20.3$ / $18.5$ & $14/10$ & $7/10$ \\
    DeltaCmd & $56.7 \pm 14.5$ & $58.0$ / $55.5$ & $23/10$ & $9/10$ & $17.3 \pm 5.4$ & $13.1$ / $21.5$ & $11/10$ & $8/10$ \\
    Teleop & $52.9 \pm 14.2$ & $43.9$ / $61.9$ & $21/10$ & $8/10$ & $15.5 \pm 5.7$ & $12.2$ / $18.9$ & $12/10$ & $\textbf{10/10}$ \\
    \textbf{Relative (Ours)} & $\textbf{42.8} \pm \textbf{5.0}$ & $\textbf{40.0}$ / $\textbf{45.6}$ & $\textbf{17/10}$ & $\textbf{1/10}$ & $\textbf{14.4} \pm \textbf{4.7}$ & $\textbf{11.0}$ / $\textbf{17.9}$ & $\textbf{11/10}$ & $8/10$ \\
    \bottomrule
    \end{tabular}
    }
\end{table*}

%% file: sections/conclusion.tex
\label{sec:conclusion}

\textbf{Conclusion.} In this work, we presented a seamless interventional method tailored for bimanual high-DoF dexterous VLA policies to address brittle deployment and compounding errors. We introduced an optimization-based relative hand retargeting method and a velocity-based shared arm control interface. Extensive real-world experiments demonstrated that our method reduces command discontinuity by up to two orders of magnitude, preserving robust manipulation capabilities post-intervention. Furthermore, we verified the efficacy of high-dimensional intervention data: the on-policy corrections collected via our Copilot shared-control act as a highly efficient patch for OOD states. By blending this data, we significantly improved the sub-goal completion rates of complex, long-horizon tasks over standard teleoperation and full takeover baselines.

\textbf{Limitations and Future Work.} Despite these advances, two main limitations remain. First, our current post-training strategy relies on simple supervised fine-tuning. Since human intervention data inherently contains sub-optimal recovery trajectories or panic-induced noise, how to more efficiently utilize this data remains an open question. Future work could explore automated data filtering or preference learning methods to extract more robust learning signals. Second, the policy's capability in extremely high-precision manipulation—such as millimeter-level drill bit alignment—remains limited. This bottleneck is primarily due to visual occlusions and the spatial resolution limits of vision-only VLA architectures, highlighting the need to integrate multi-modal sensing in future iterations.